\documentclass[a4paper,12pt]{article}
\usepackage{tabularx} % extra features for tabular environment
\usepackage{mathtools}
\usepackage{cases}
\usepackage{amsmath}  % improve math presentation
\usepackage{graphicx} % takes care of graphic including machinery
\usepackage[margin=1in]{geometry} % decreases margins
\usepackage[final]{hyperref} % adds hyper links inside the generated pdf file
\usepackage{epstopdf}
\usepackage{caption}
\usepackage{subcaption}
\usepackage{float}

\usepackage[table,xcdraw]{xcolor}

\usepackage[backend=biber,style=ieee]{biblatex}
\usepackage[affil-it]{authblk}
\usepackage{lmodern}

\title{Explanations of Machine Learning predictions: a mandatory step for its application to Operational Processes}

\author[1,2]{Giorgio Visani\thanks{Electronic address: \texttt{giorgio.visani2@unibo.it}; Corresponding author}}
\author[1]{Federico Chesani}

\author[2]{Enrico Bagli}
\author[2]{Davide Capuzzo}
\author[2]{Alessandro Poluzzi}

\affil[1]{Università degli Studi di Bologna, Dipartimento di Ingegneria e Scienze Informatiche, viale Risorgimento 2, 40136 Bologna (BO), Italy}

\affil[2]{CRIF S.p.A., via Mario Fantin 1-3, 40131 Bologna (BO), Italy}

\makeatletter
\def\@maketitle{%
  \newpage
  \null
  \vskip 2em%
  \begin{center}%
  \let \footnote \thanks
    {\LARGE\bfseries \@title \par}%
    \vskip 3em%
    {\normalsize
      \lineskip .5em%
      \begin{tabular}[t]{c}%
        \@author
      \end{tabular}\par}%
    \vskip 2em%
    {\normalsize \@date}%
  \end{center}%
  \par
  \vskip 3em}
\makeatother
\date{\parbox{\linewidth}{\centering%
  August 5, 2019\endgraf\bigskip\medskip
  Presented at:\endgraf\smallskip
  $16^{th}$ Credit Scoring and Credit Control Conference (CRC)\endgraf
  28-30 August 2019, Edinburgh, UK }}

\hypersetup{
	colorlinks=true,       % false: boxed links; true: colored links
	linkcolor=blue,        % color of internal links
	citecolor=blue,        % color of links to bibliography
	filecolor=magenta,     % color of file links
	urlcolor=blue         
}
\begin{document}

\maketitle

\begin{abstract}
In the global economy, credit companies play a central role in economic development, through their activity of money lenders. This important task comes with some drawbacks, mainly the risk of the debtors of not being able to repay the provided credit. Therefore, Credit Risk Modelling (CRM), namely the evaluation of the probability that a debtor will not repay the due amount, plays a paramount role. Statistical approaches have been successfully exploited since long, becoming the most used methods for CRM. Recently, also machine and deep learning techniques have been applied to the CRM task, showing an important increase in prediction quality and performances. However, such techniques usually do not provide reliable explanations for the scores they come up with. As a consequence, many machine and deep learning techniques fail to comply with western countries regulations such as, for example, GDPR. In this paper we suggest to use LIME (Local Interpretable Model-agnostic Explanations) technique to tackle the explainability problem in this field, we show its employment on a real credit-risk dataset and eventually discuss its soundness and the necessary improvements to guarantee its adoption and compliance with the task.
\end{abstract}

\pagebreak

\section{Introduction}
Operational Processes are defined as the core business of companies and firms: drug companies consider them to be drug testing and approval, manufacturing firms identify them in the product assembly process, while banks and financial firms have their own core business in risk management and evaluation.
\medskip 

In order to be able to concede loans, financial institutions are compelled to predict whether an applicant is likely to repay the debit. In such a framework, Credit Scoring plays a huge role in ranking applicants based on their likelihood to pay back the loan. Each person is associated with a credit score value, namely a “number that summarizes its credit risk, based on a snapshot of its credit report at a particular point in time” \cite{noauthor_fdic:_2007}. Behind the scenes, CRM is employed to reach the goal: scoring models, or “scorecards”, are generated from historical data, employing well-established statistical techniques.
\medskip

The cornerstones of a reliable scorecard are well depicted by Loretta Mester in \cite{mester_what_1997}: “the model should give a higher percentage of high scores to borrowers whose loans will perform well and a higher percentage of low scores to borrowers whose loans won’t perform well”. Several advantages stem from risk modelling, among the most important there are an increased profitability of financial corporations due to more reliable loans conceded, the chance of evaluating new loan programs based on the data collected and the enhancement of the credit-loss management capability \cite{noauthor_fdic:_2007}. Therefore, over the years, some institutions arose to accomplish the task.
\medskip

CRIF is a global company specialized in credit bureau and business information, outsourcing and processing services, and credit solutions. Its expertise in CRM dates back to the 80s, making the company one of the leaders in the Italian CRM market as well as an important benchmark worldwide. Nowadays, one of CRIF’s endeavours is towards the adoption of advanced analytics for CRM.
\medskip

This work represents a joint effort between CRIF and the University of Bologna. The aim is to exploit the state-of-the-art Machine Learning techniques and gain the benefits of their higher accuracy, while retaining the ability of producing reliable explanations about the models’ output. We consider it the first step to make such models adherent to the GDPR standards of UE and the more demanding countries.
\medskip

In this contribution, we introduce our approach, where classical (statistical) models are exploited alongside with Machine Learning ones, thus taking advantage of the enhanced accuracy. Then, we employ ground-breaking techniques, i.e. LIME, to achieve some interpretability of the outcomes. Prediction techniques and explanation capabilities are evaluated on a Credit Risk dataset. Eventually, we illustrate the approach by grounding it on few examples, showing the provided explanations, and discussing its reliability.

\section{The Models}

\subsection{Classical Credit Risk Models}
CRM is a relatively long-established procedure, since models were employed to solve the task from the 1950s. The classical methodology is based on a variety of techniques stemming from the statistics field, the most popular ones are Linear, Logistic and Probit models.
\medskip

The three of them fall inside the wider class of prediction models, i.e. mathematical models that aim to predict the values of a target variable, by knowing only the values of some auxiliary variables. The target variable represents the event or quantity of interest. The aim is to find a relation that links the target variable with the auxiliary variables, also called regressors or independent variables.  The knowledge of such dependence can be exploited in order to predict the value of the target variable before the event has happened.
\medskip

In credit risk, the target variable is the default of the borrower person. It is usually coded as 1 if the default occurred, 0 otherwise. Since only two values are allowed, the variable is said to be binary.  In order to predict it, statistical models consider the probability of default (from now on PD), which can assume any continuous value from 0 to 1.
\medskip
 
Following the parametric approach, the relation in mean between PD and the regressors is considered and a guess is made about its functional form. The functional form is the mathematical formulation of a function which can be drawn in the space of the independent variables; tweaking its parameters the shape of the function will be modified. Once the shape of the relation has been chosen, the parameters are inferred from the historical data, with the goal of drawing the closest function to the real relation between PD and the regressors.
\medskip

Below, the functional form underlying the three most-used models in Credit Risk:
\begin{numcases}{\text{PD}(x) = \Pr(Y=1|X=x) =}
\mathbf{X}^T \beta^{(1)} & \text{Linear Model}  \label{eq:1}\\
& \nonumber \\
\frac{\exp(\mathbf{X}^T \beta^{(2)})}{1+\exp(\mathbf{X}^T \beta^{(2)})} \qquad &\text{Logistic Model} \label{eq:2}\\
& \nonumber \\
\Phi(\mathbf{X}^T \beta^{(3)})  &\text{Probit Model} \label{eq:3}
\end{numcases}
, where $\mathbf{X}$ represents the matrix of independent variables, while $x$ stands for a particular realization of the multivariate random variable $X$. \\
$\beta^{(1)}, \beta^{(2)}, \beta^{(3)}$ are the parameters respectively of the Linear, Logistic and Probit Models. \\ $\Phi(\cdot)$ is the Cumulative Distribution Function of a standard Gaussian, $\mathit{N}(0,1)$.
\bigskip

Linear models (\ref{eq:1}) assume a linear relation, which can be represented as a hyperplane, in the space of regressors. Its main strength is the ease in estimation and explanation, although the linearity is too strict assumption in many cases. Moreover, the predictions of linear models are not bounded and can assume values ranging over all the real numbers; when modelling a probability, as in the Credit Risk case, this is a major drawback since the true probabilities range from 0 to 1.
\medskip

\begin{figure}[h]
\centering
\includegraphics[width=0.5\textwidth]{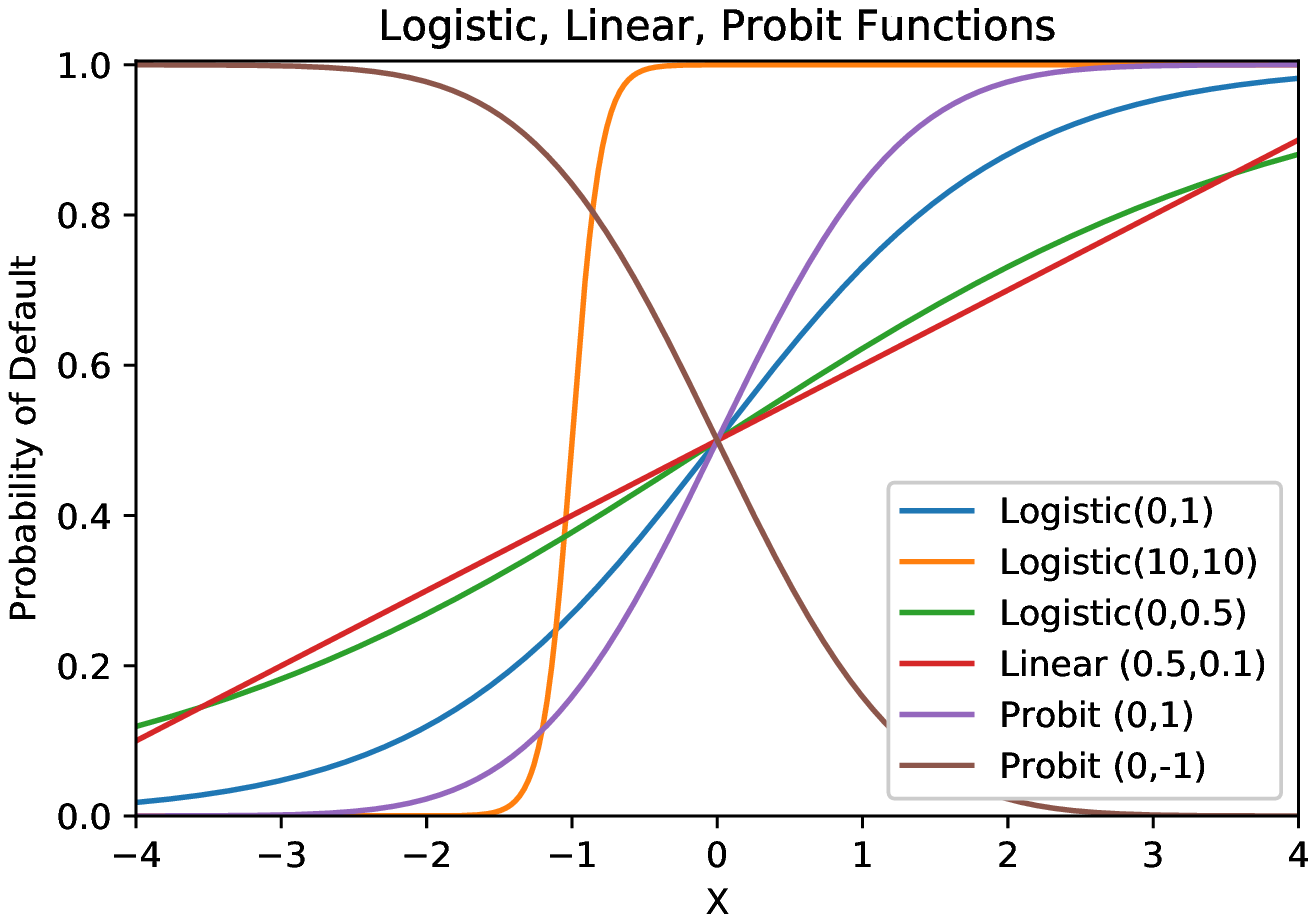}
\captionsetup{format=hang}
\caption{Shape of Logistic, Probit and Linear functions, associated with different parametrization. \\
 The Probability of Default, ranging from 0 to 1, is modelled against a single independent variable X.}
\label{logit_probit_linear_functions}
\end{figure}

To overcome those issues, Logistic (\ref{eq:2}) and Probit (\ref{eq:3}) regressions are quite useful. Both of them transform the probability into a new variable spanning over the entire real line. The transformation is a bijective function, meaning that it is always possible to convert each value of the new variable back into the probability value that generated it. In doing so, both Logistic and Probit have the additional advantage of modelling the relation in a non-linear way: it is possible to draw a curve on the independent variables’ space. It is a dramatic increase in the representation capability, even if such curvy lines are bound to be monotically increasing or decreasing (as shown in Figure \ref{logit_probit_linear_functions}).

An additional perk of Logistic Regression, when compared to Probit, is its interpretability of results: the parameters derived from the best curve’s estimation, can be regarded as odds ratio, i.e. the ratio between the probability of default and non-default, namely $\frac{P(Y=1|X=x)}{P(Y=0|X=x)}$.
\smallskip

Starting from the mean value of one specific independent variable, the increase of 1 unit brings an increase of the odds ratio that is equal to the exponentiated parameter. The relation is valid when the other regressors' values are fixed to their mean.
\par 
This benefit is due to the particular transformation employed by the model, which preserves the chance of interpreting the results.

\subsection{Machine Learning Models}

The classical definition of Machine Learning dates back to 1997 on behalf of Tom Mitchell \cite{mitchell_tom_machine_1997}: 
\begin{quote}
A computer program is said to learn from experience E with respect to some class of tasks T and performance measure P if its performance at tasks in T, as measured by P, improves with experience E.
\end{quote}
\medskip

By this train of thought, almost any kind of prediction algorithm may fall into the class of Machine Learning models. Consider Logistic Regression, the parameter tuning phase is done through an iterative algorithm, usually Newton-Raphson, which improves the estimated model’s performance at each iteration, measured by the increase in the Likelihood value.
\medskip

Because of the extent of such a general framework, in this contribution we consider only non-parametric Machine Learning models. They estimate the relationship between the target variable and the predictor variables, without constraining it to have a precise functional form. This peculiarity allows to model non-linear relations of any possible shape, making the technique more flexible compared to classical parametric methods. Machine Learning models of this kind usually outperform classical methods in non-linear settings and achieve the same results when the nature of true relations is simply linear.
\bigskip

We decided to focus on tree-based Machine Learning models, in particular Gradient Boosting Trees. This is because they retain the enhanced predictive power of Machine Learning models, while having the additional advantage of requiring almost none pre-processing. Because of their structure, they are able to cope with outliers and extreme values easily. Besides, theoretically they are also able to address missing values.
Unfortunately, dealing with missing values can be computationally unfeasible, so that the vast majority of the method's implementations do not allow such peculiarity.
\medskip

Broadly speaking, Gradient Boosting Models rely on the idea of creating many simple and weak models, also called learners, and to aggregate them sequentially into an Ensemble Model.
\medskip

In the case of Gradient Boosting Tree Model, Single Decision Trees are employed as weak learners. Each Tree is grown on the same dataset, slightly modified at each step: a different weight is given to each unit, based on the prediction error of the ensemble model built so far. Thereby, units which are already predicted well are given low weights, whereas individuals presenting imprecise or wrong predictions will benefit of higher weights. This allows the following trees to focus more on the hard to predict individuals.
\medskip

On one hand, this approach allows to create an ensemble model, able to predict well also complex and highly non-linear parts of the regressors’ hyperspace. On the other hand, this kind of structure is prone to overfitting and it requires cross validation and checks on the performance out-of-the-box.
\medskip

Weak Learners refer to very simple models which achieve modest performance (usually their accuracy is just above chance), such as Single Decision Trees with very few branches. They are employed in boosting procedure, instead of strong learners, because this helps the algorithm to learn “slowly”: small performance improvements are made per each weak Tree added to the Boosted Model. It allows to prevent overfitting and to give the chance to the Ensemble to learn different paths to predict well the same region. Doing so, increases the robustness of the final model and helps also to keep it simple, e.g. when Decision Trees with just one split are employed, namely stumps, the Boosting final model can be regarded as an additive model. \cite{james_introduction_2013}
\medskip

The best single tree to be added at each step, is chosen minimizing the loss function that compares the true values of the response variable and the predictions of the boosted ensemble until the present step. The gradient of the loss function is calculated with respect to the parameters' random variables and the result is a vector in the parameters hyperspace. Such vector contains the necessary information to retrieve the next tree formulation, namely the one that guarantee the greatest improvement on the loss function, hence the name Gradient Boosting. 
\par
\begin{figure}[h!]
\centering
\includegraphics[width=1\textwidth]{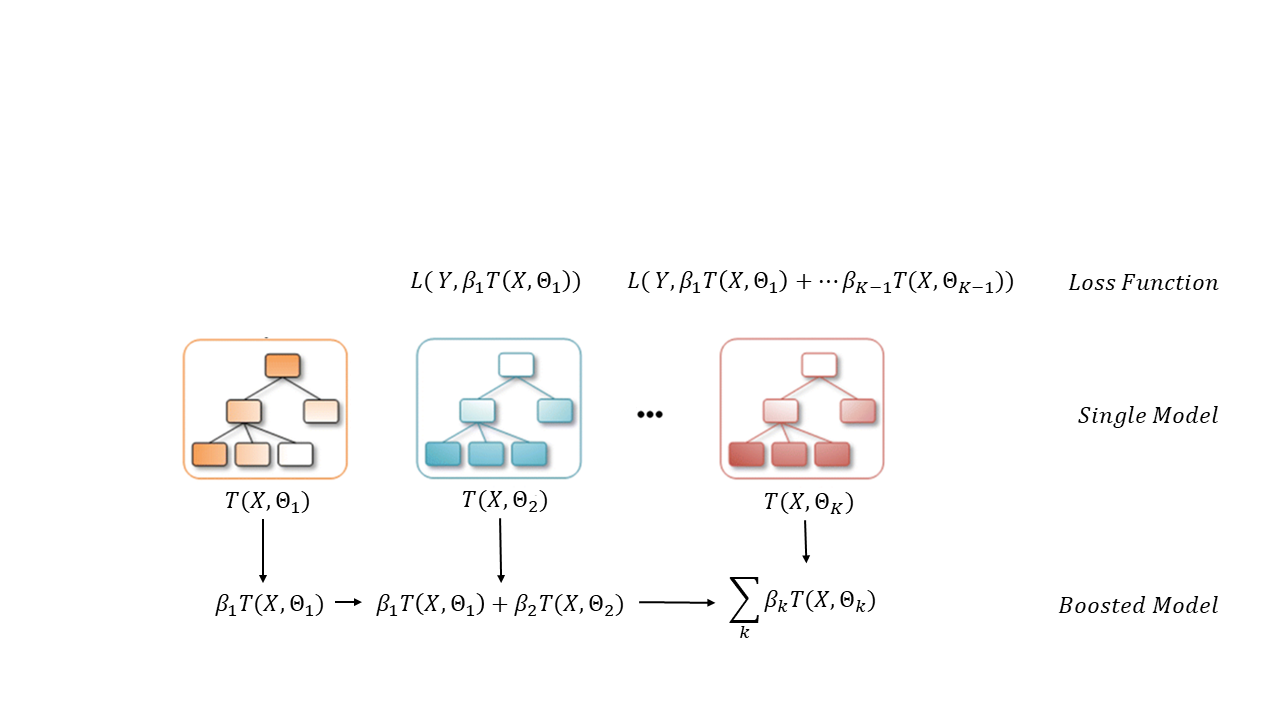}
\captionsetup{format=hang}
\caption{Gradient Boosting Tree Model construction. 
\vspace{0.5em} \\
$T(X,\Theta_k)$ is the best Tree built at step $k$, its parameters $\Theta_k$ are chosen in order to minimize the Loss Function between the target variable $Y$ and the Boosted Model of the previous step. \\
The $\beta_k$ parameter is the weight of the Tree, when added in the Boosted Ensemble, this is also chosen with respect to the Loss Function.}
\label{gbm}
\end{figure}

 
The entire framework of Boosted Tree Models is thoroughly explained in Figure \ref{gbm}.
\bigskip

Gradient Boosting Models can be implemented using a wide range of different single classifiers, although the most widespread and good performing architecture employs Single Trees. Due to its pervasive use, it enjoys very fast and reliable implementations: XGBoost \cite{noauthor_xgboost_nodate}, LightGBM \cite{noauthor_welcome_nodate}, CatBoost \cite{noauthor_catboost_nodate} libraries are among the well-known ones. 
\medskip

\section{Model Performance Analysis}
The dataset used in this paper for the performance analysis comes from an anonymized statistical sample, representative of an application process. It has been obtained by pooling data from several Italian financial institutions.
\medskip

The definition of “bad payer” changes case-to-case. In the present application, bad payers were considered both: users with 90 or more days past due for at least one payment towards the bank, or individuals with at least one shift from a past due to an actual loss in the last 12 months. The composition of the dataset is shown in the Table \ref{tab:1}. 
\medskip

In order to test the consistency of a model, it is good practice to split the dataset observations into two non-overlapping samples. The Training Set, consisting in 70\% of the entire dataset, is employed to tune the algorithm; whereas the Test Set, composed of the remaining 30\% of the observations, is useful for checking the algorithm’s performances on new data. This allows to have reliable values of the chosen evaluation metric: the figures obtained on the Test Set will be similar to the ones achieved by the algorithm when predicting brand-new individuals. The sampling has been made at random, to be sure that the model will not capture situational patterns.
\medskip

\pagebreak

Here are the effective dimensions of the two data sets:
\bigskip

\begin{table}[h]
\centering
\begin{tabular}{|c|c|c|}
\cline{1-3}
\textbf{Data set name} & \textbf{Population} & \textbf{\%Bad} \\ \cline{1-3}
Training set           & 39.418              & 2,9\%  \\ \cline{1-3}
Test set               & 16.893              & 3,1\%  \\ \cline{1-3}
Total                  & 56.311              & 3\%   \\ \cline{1-3}
\end{tabular}\\
\captionsetup{format=hang}
\caption{Dataset Composition \\
The Train-Test split has been done in a balanced fashion: the ratio between number of non-compliant individuals and total number of applicants is statistically non different
}
\label{tab:1}
\end{table}
\medskip

The most reliable figure of merit of the model performances in CRM field is the Gini Index \cite{hand_modelling_2001}.
\medskip

On the previously discussed dataset, we applied both the classical techniques, embodied by Logistic Regression, as well as the Gradient Boosting one, belonging to the class of newly devised Machine Learning models. The development of the two architectures has been kept completely separate, based on the same choices on the dataset, in particular the Train-Test split.
\medskip

\begin{figure}[h]
\centering
\includegraphics[width=0.5\textwidth]{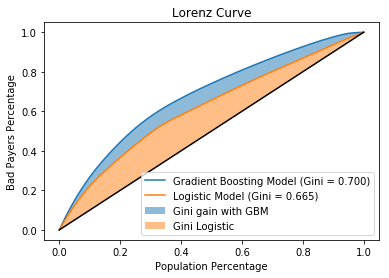}
\captionsetup{format=hang}
\caption{Lorenz Curve Comparison\\
Gradient Boosting vs Logistic Regression}
\label{lorenz_curve}
\end{figure}

In the Figure \ref{lorenz_curve} we compare the two models on the Gini index achieved. It is possible to recognize an improvement in performance, testified by a Gini increase of more than 3 points. 

\section{Explaining the Machine Learning Model: LIME}
Despite the enhanced accuracy, Machine Learning Models display weakness especially when it comes to interpretability. In order to address it, we approached the problem employing state-of-the-art techniques, in particular LIME. LIME is a method for explaining Machine Learning Models that behave like black-boxes, developed by Marco Tullio Ribeiro in 2016 \cite{ribeiro_why_2016}.
\medskip

\begin{figure}[h]
\centering
\includegraphics[width=0.5\textwidth]{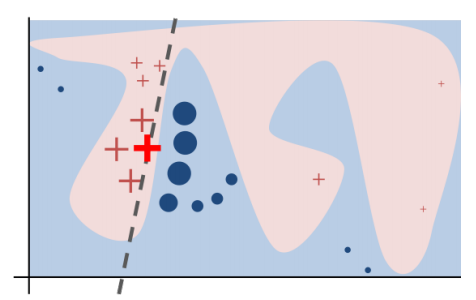}
\captionsetup{format=hang}
\caption{LIME’s modus operandi. \\
Courtesy of Marco Tullio Ribeiro \cite{ribeiro_why_2016}}
\label{LIME}
\end{figure}

Chosen a given individual and the relative prediction made by the black-box model, LIME returns the most important variables that drove the model towards that particular decision. It can be made for each individual (Local Explanation). The main idea is to consider the space of dataset variables, the black-box model can be thought of as a plane in such space, dividing good from bad payers, as shown in Figure \ref{LIME}. Each point in the space represents a person in the dataset.
\medskip

When an individual is chosen, LIME generates fictitious points close to it, i.e. they show similar values of the variables. It predicts the behaviour of the generated individuals using the black-box model. The model predictions on the new points become the response variable and LIME predicts it using a Linear Model, specifically Ridge Regression in order to prevent overfitting \cite{hoerl_ridge_1970}. This is achieved thanks to the regularization term, namely the sum of squared coefficients (squared $\ell 2$ norm), inside the Loss Function:

\begin{equation}\label{eq:4}
\arg\min_{\beta \in \mathbf{R}^p} \quad (Y_{ML}-\mathbf{X}^T \beta)^T (Y_{ML}-\mathbf{X}^T \beta) + \lambda ||\beta^T \beta ||_{2}^2
\end{equation}

, $Y_{ML}$ represents the Machine Learning model's predictions per each individual, which is now regarded as the response variable.
\medskip 

When fitting the Linear Model, the best variables’ coefficients are chosen according to Equation (\ref{eq:4}). It is therefore possible to retrieve the most important predictors, namely the ones with the highest coefficients (in absolute terms). The highest coefficients represent LIME’s model explanation, since they describe the strength of the variables’ impact on the Default Probability inferred by the Machine Learning model.
\bigskip

We tested LIME on several data points. In Table \ref{tab:2}, we show LIME interpretations for one “good” user (on the left) and one “bad” (on the right). The three graphs come from three separate employments of LIME on the same unit and the same Gradient Boosting Model. This is to check whether the explanations are stable, namely if different calls to LIME return equivalent results. Our application obtains satisfactory stability, as the graphs for the same unit are similar, except for small changes in variables' magnitude due to LIME's random sampling. 
\medskip

The sum of the bars' values, along with the intercept, produces the Local Ridge Model prediction (in Table \ref{tab:2} denoted as LIME Prediction). The bars' length highlight the specific contribution of each variable: the green ones push the model towards "good payer" prediction, whereas the red ones to "bad payer".
\pagebreak

%TABLE CON LE 6 EXPLANATIONS DI LIME
\begin{table}
\begin{tabular}{|c|c|}
\hline
\rowcolor[HTML]{96FFFB} 
Unit Number: 1  &Unit Number: 53\\ 
\rowcolor[HTML]{96FFFB} 
GBM Prediction: 0.059  &GBM Prediction: 0.956  \\ 
\rowcolor[HTML]{96FFFB} 
LIME Prediction: 0.054  &LIME Prediction: 1.031 \\ 
\rowcolor[HTML]{96FFFB} 
Model $R^2$: 0.738  &Model $R^2$: 0.506 \\
\rowcolor[HTML]{96FFFB} 
Intercept: 1.104  &Intercept: 0.424 \\  \hline
& \\
\centering
 \includegraphics[width=0.45\linewidth]{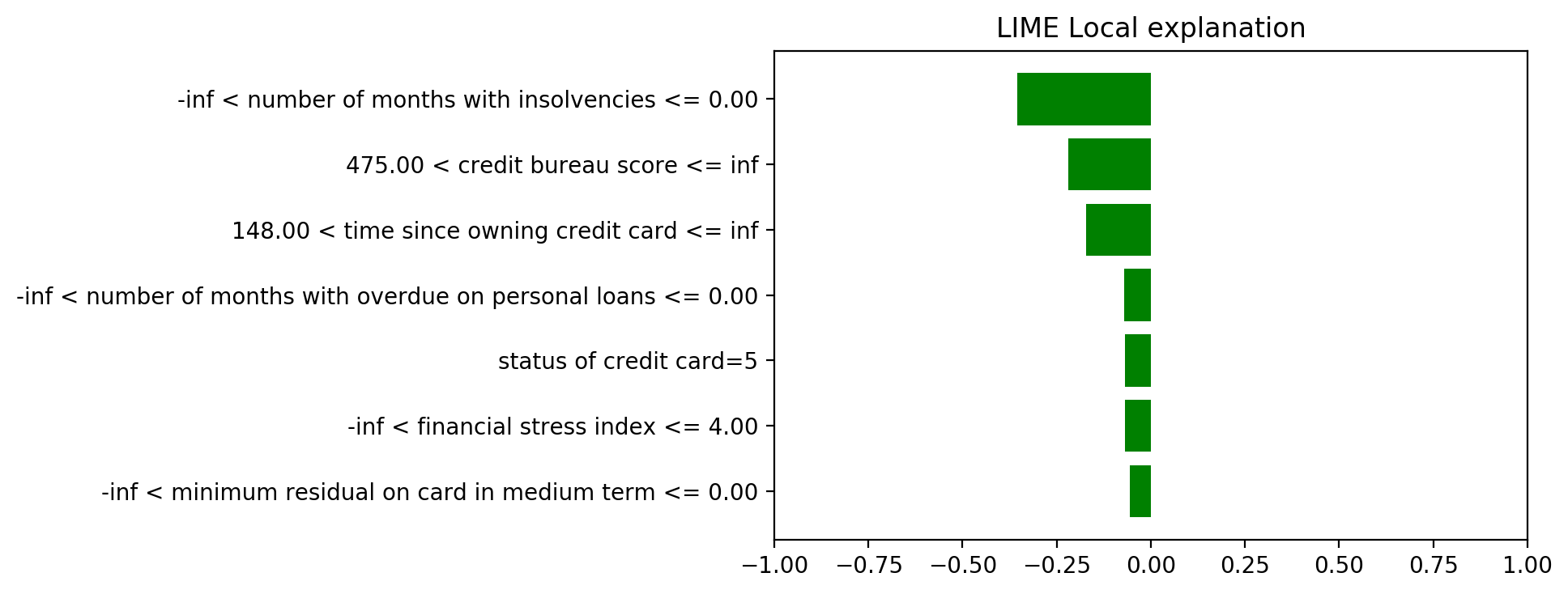}& \includegraphics[width=0.45\linewidth]{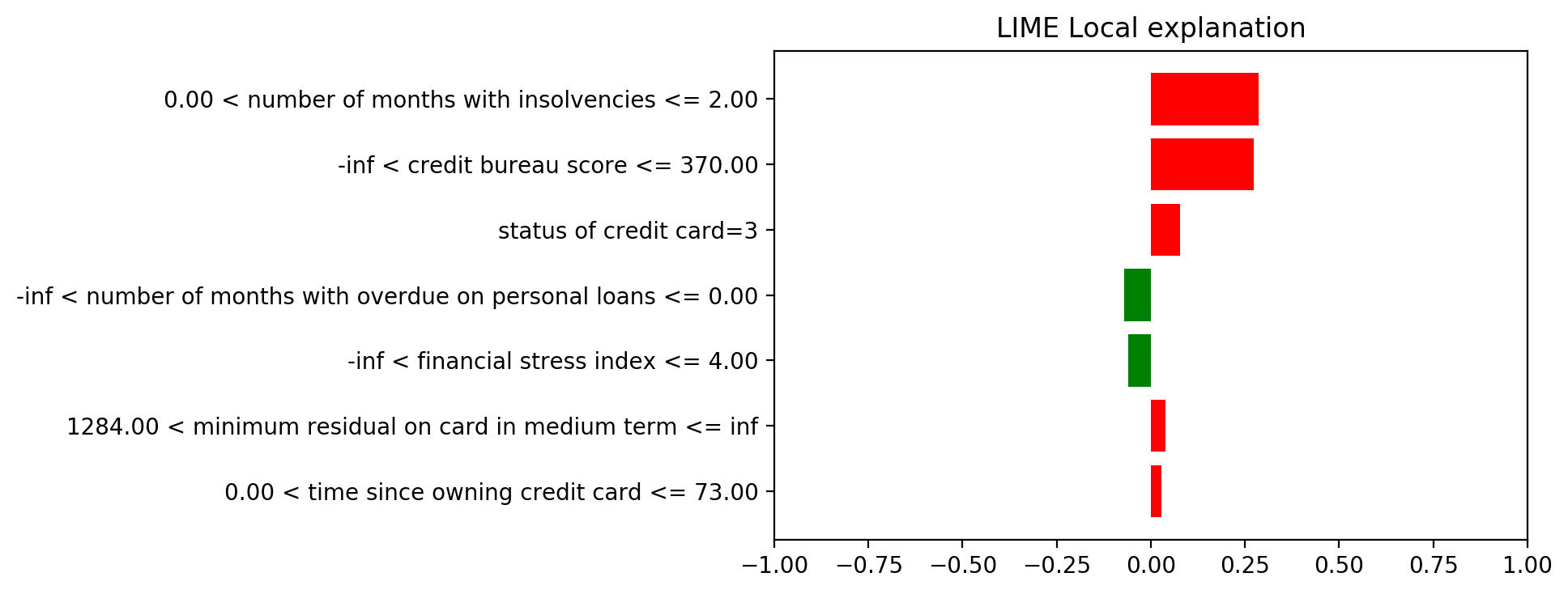} \\ \hline
& \\
 \includegraphics[width=0.45\linewidth]{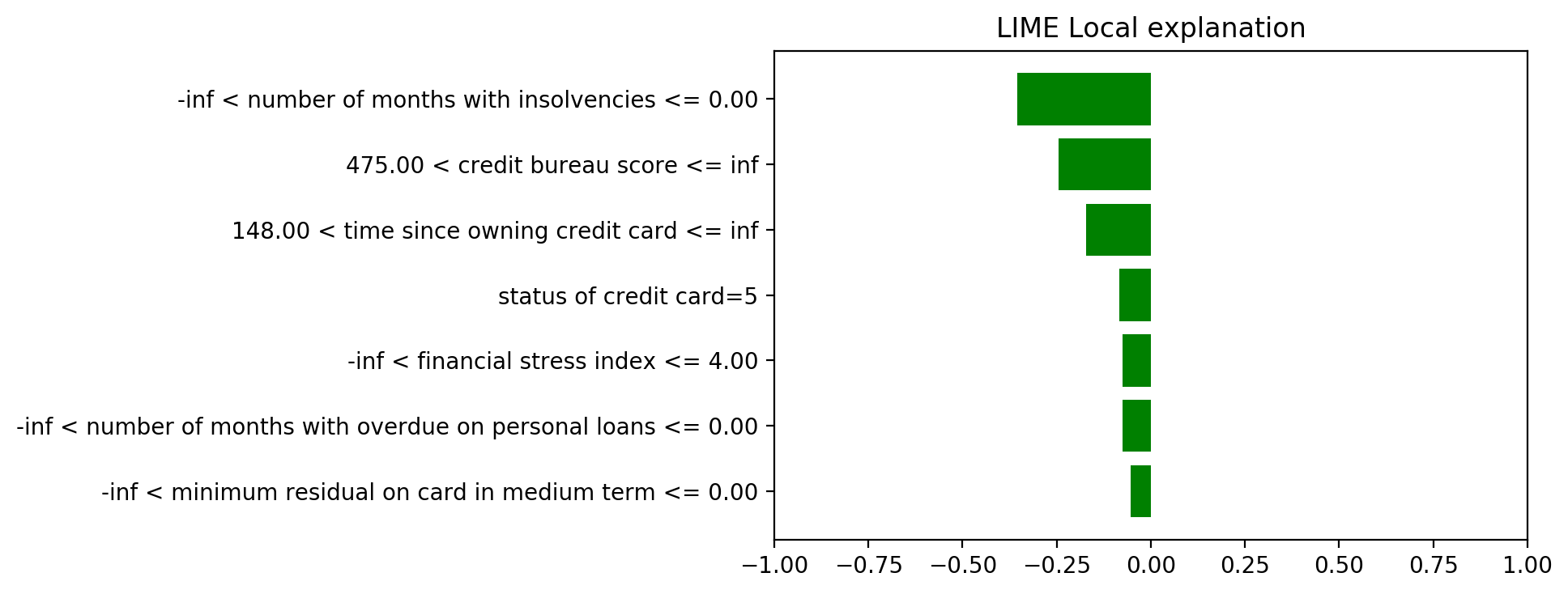}& \includegraphics[width=0.45\linewidth]{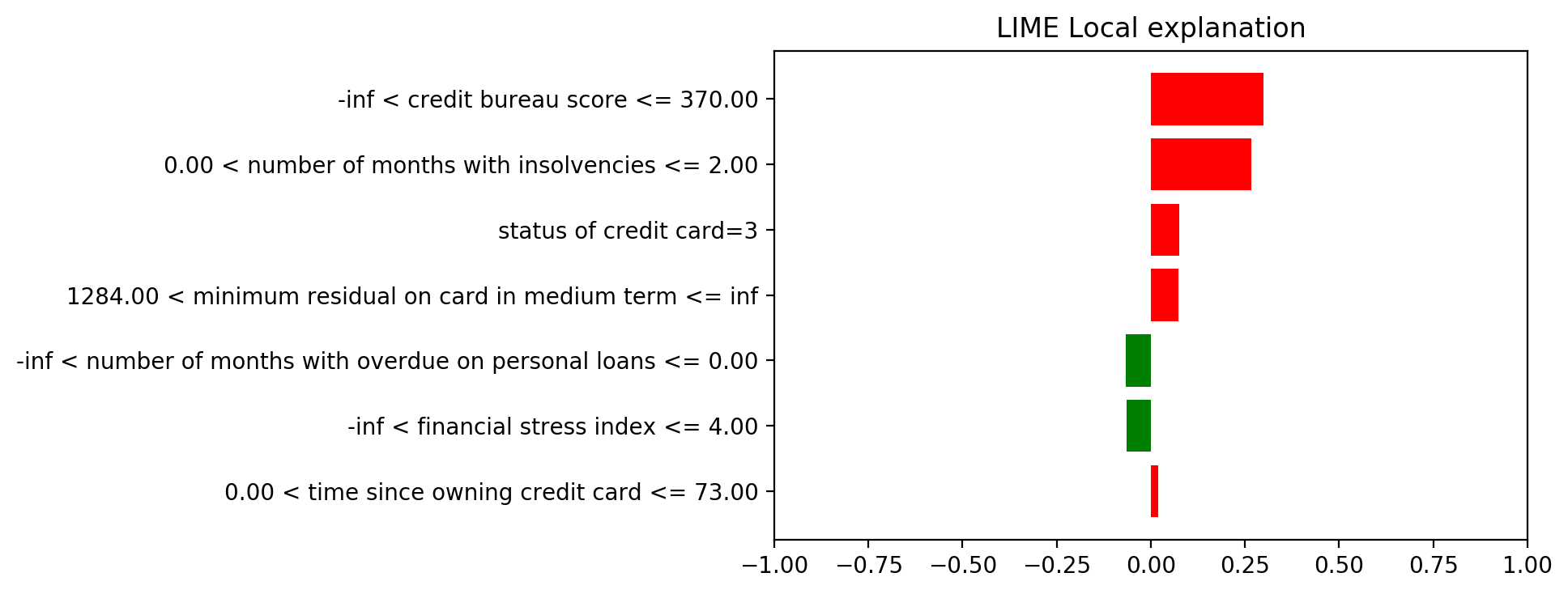} \\
\hline
& \\
 \includegraphics[width=0.45\linewidth]{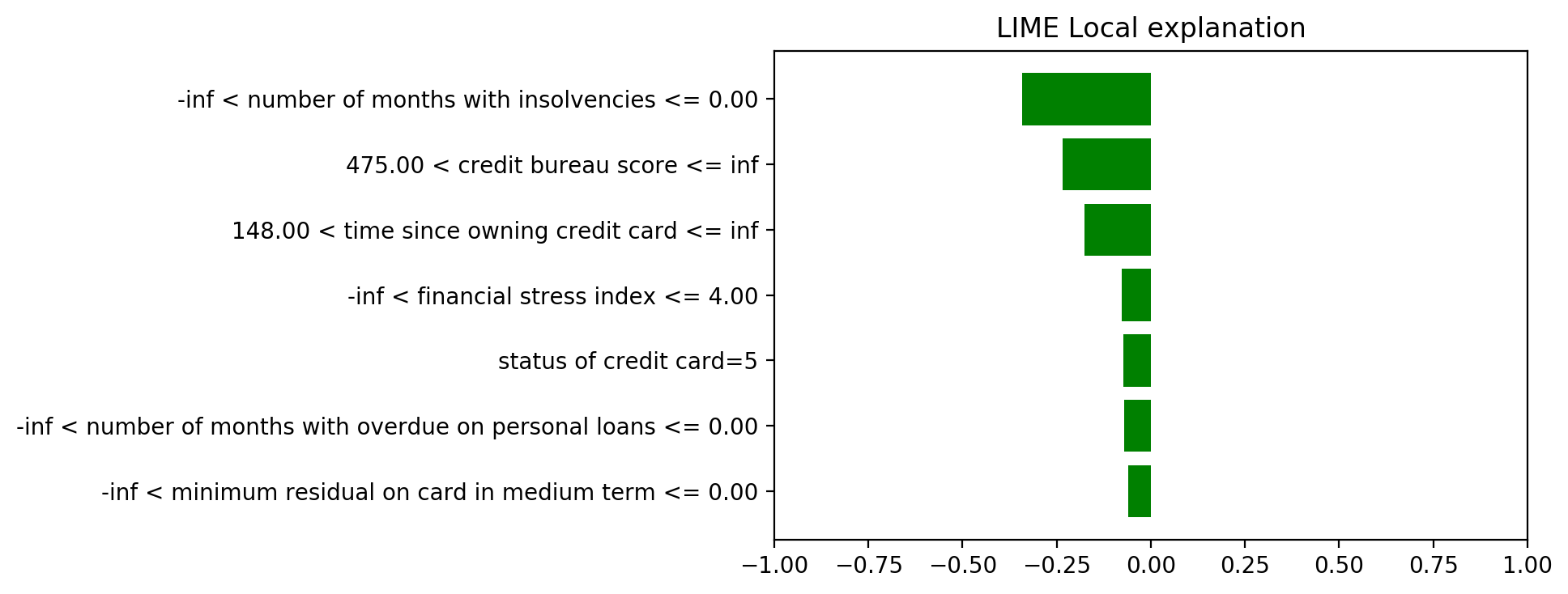}& \includegraphics[width=0.45\linewidth]{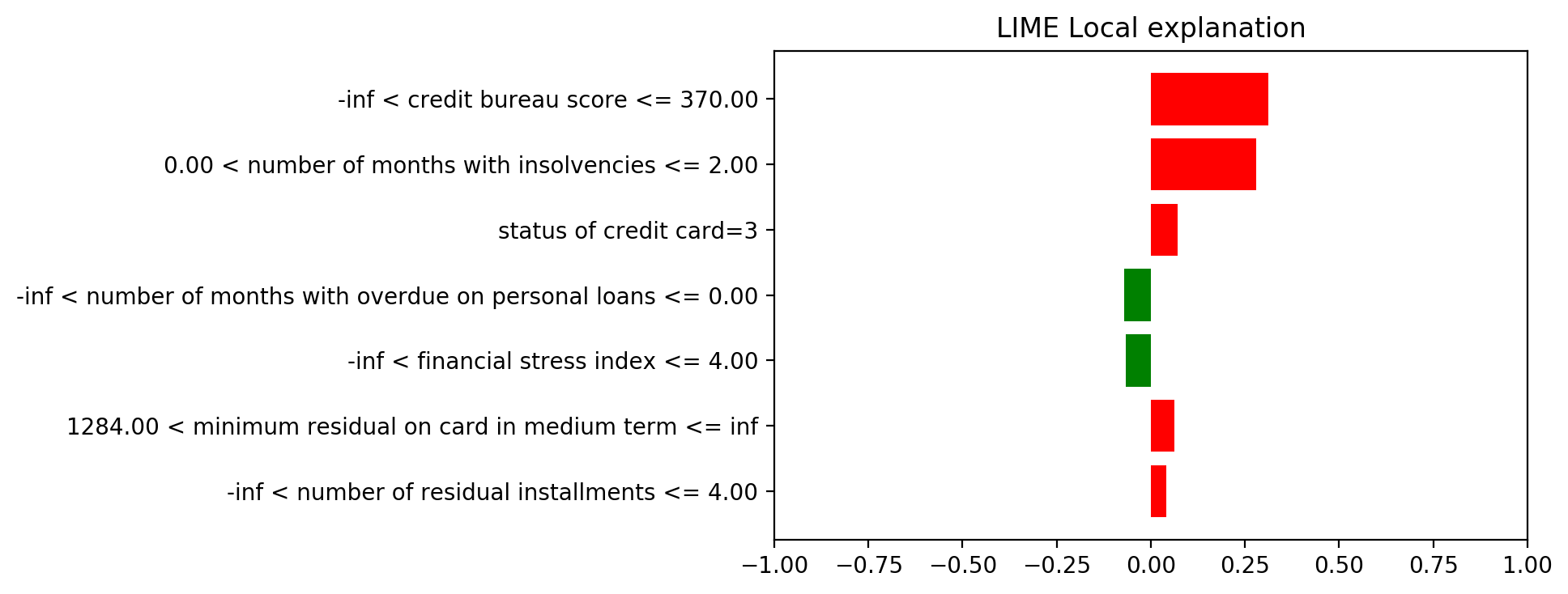} \\ \hline
 \end{tabular}
 \captionsetup{format=hang}
 \captionof{table}{Examples of LIME Stability.
 \vspace{0.5em} \\
On the left, there are 3 different applications of LIME on the same unit (classified as Good Payer by the Gradient Boosting Model); on the right the same idea applied to a unit classified as Bad Payer. In these settings LIME Explanations are stable.} \label{tab:2}
\end{table}

We show only the seven more important variables in order to explain the Gradient Boosting Model. In the two use-cases, the key regressors are the Credit Bureau Score (CBS), namely a comprehensive value developed using information provided by the Italian Credit Bureau, and the number of months where unpaid installments occurred, within the last year.
\par
On the left part, the user exhibits 0 months with unpaid installments and falls inside a good class of CBS index.
Such circumstances are the major ones leading Gradient Boosting Model to classify him as a good payer.
\par
On the contrary, the user shown on the right displays at least one month with insolvencies and he falls inside a bad CBS class, these conditions drive the model to classify him as bad payer.
\medskip

An interesting aspect is represented by LIME's prediction value of the "bad" individual that is more than 1, although, as a probability value, it should be at most 1. This behaviour stems from the Ridge Regression Model employed: since it is a particular case of Linear Model, it does not bound the predictions in the interval $[0,1]$.
\par 
The interpretation provided by LIME seems plausible and especially valuable for any institute, which requires to be able to communicate the reason for granting or rejecting the loans.
\bigskip

However, “all that glisters is not gold”: LIME shows also some weak points, we highlight the two major ones \cite{molnar_interpretable_nodate}.
\medskip

\begin{figure}[H]
 \centering
\begin{subfigure}[t]{0.4\textwidth}
\centering\includegraphics[width=0.90\textwidth]{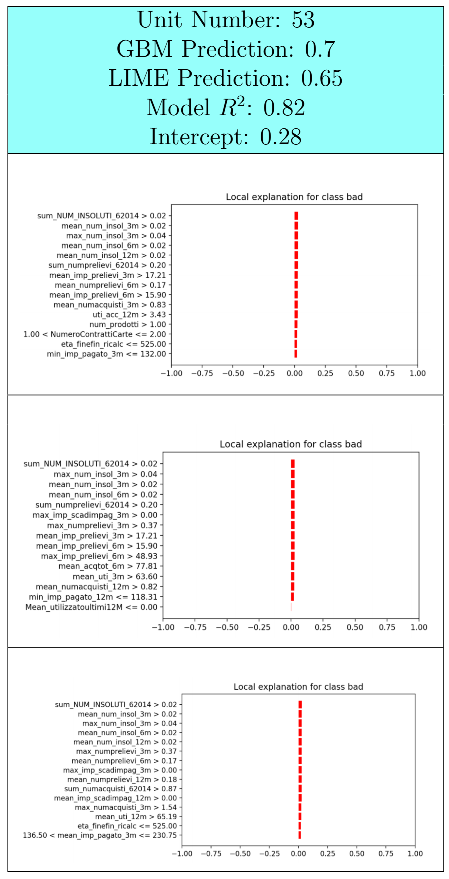}
\captionsetup{format=hang}
\caption{LIME explanations are not informative
when applied to Machine Learning
Models with many
independent variables, in this case 100.}
\label{LIME_cattivo1}
\end{subfigure}
\begin{subfigure}[t]{0.5\textwidth}
\centering
\includegraphics[width=0.95\textwidth]{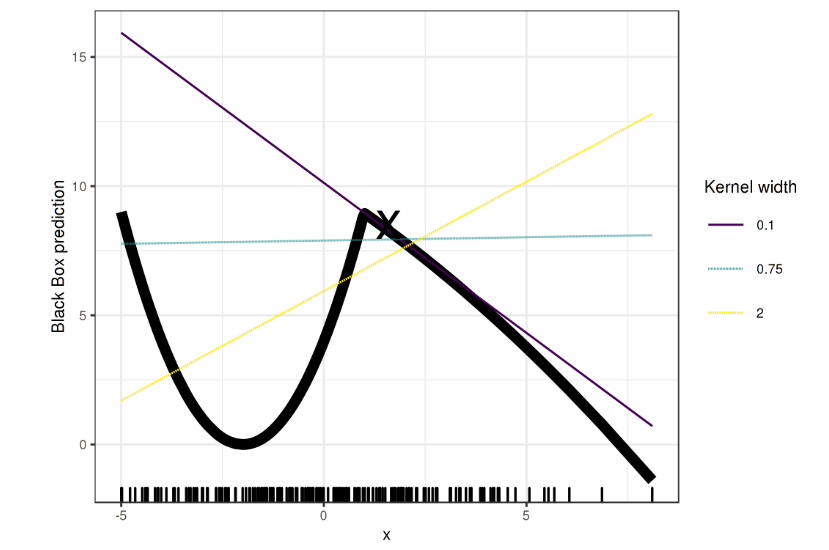}
\captionsetup{format=hang}
\caption{LIME attempts of drawing a straight line on a non-differentiable point of the curve. \\
This results in great sensitivity to parameters tweaking (in particular kernel width). \\
Courtesy of Christoph Molnar in \cite{molnar_interpretable_nodate}.}
\label{LIME_cattivo2}
\end{subfigure}
\caption{LIME Issues}
\label{LIME_cattivo}
\end{figure}

Firstly, LIME is sensitive to dataset dimensionality: huge number of variables may cause the local explanation to be  unreliable as well as not to discriminate among relevant and irrelevant variables. The first issue may be spotted by a low value of $R^2$ metric and a radical change of the most important variables in distinct LIME applications. The second problem is shown in Figure \ref{LIME_cattivo1}, where the most relevant regressors exhibit low values and many of them are equally important.
\par
In addition, correlation among variables causes the method to fail: even with few but correlated predictors, the explanations become unstable. 
\par

Such weaknesses are a major drawback since the most recent models can handle big and complex datasets with correlated variables inside; when it comes to explain their results on such data, LIME does not provide reliable interpretation.
\medskip

Secondly, from practitioner’s perspective LIME might be cumbersome, it is indeed not as intuitive as it seems. There are many model assumptions below the surface of the method, embodied by a variety of parameters. While tweaking the parameters, to achieve a better explanation, it is important to keep in mind what they represent. The danger is to create a linear model that is not adherent to the Machine Learning one, because some of the assumptions do not hold for the employed dataset.
\par

In addition, even if LIME parameters have been tuned carefully, it can happen to end up with a poor linear model. There are indeed some critical points in which LIME method may be inadequate:  when a non-differentiable point on the Machine Learning surface is chosen, it is hard to find a linear model that fits well in its neighbourhood. By definition it does not exist a tangent line to the surface in that point so, it becomes hard to make linear explanations. This is the situation depicted in Figure \ref{LIME_cattivo2}.

%\begin{table}[]
%\begin{tabular}{|c|}
%\hline
%\rowcolor[HTML]{96FFFB} 
%Unit Number: 53\\ 
%\rowcolor[HTML]{96FFFB} 
%GBM Prediction: 0.7  \\ 
%\rowcolor[HTML]{96FFFB} 
%LIME Prediction: 0.65 \\ 
%\rowcolor[HTML]{96FFFB} 
%Model $R^2$: 0.82 \\
%\rowcolor[HTML]{96FFFB} 
%Intercept: 0.28 \\  \hline
%\\
%%\centering
% \includegraphics[width=0.45\linewidth]{images/LIME_cattivo_1.png} \\ \hline
%\\
% \includegraphics[width=0.45\linewidth]{images/LIME_cattivo_2.png} \\
%\hline
%\\
% \includegraphics[width=0.45\linewidth]{images/LIME_cattivo_3.png} 
% \\ \hline
%\end{tabular}
%\end{table}

\section{Discussion and conclusions}
As highlighted in the use case above, Gradient Boosting gives rise to higher Gini Index compared to Logistic Regression, thanks to the chance of incorporating non-linear trends into the model. This result shows the main benefit of using Machine Learning models for Credit Risk Modelling, since they give better predictions compared to classical ones. Another benefit of some Machine Learning models, especially the ones based on Decision Trees, is that they require very little data pre-processing, resulting in faster and less error-prone Score creation.
\medskip

Regrettably, to date there is no methodology allowing unambiguous explanations of Machine Learning models. Such explainability issue has held back their adoption in the CRM financial field. A convincing solution to this problem is, therefore, a mandatory step on the path towards Machine Learning models employment, in accordance with the existing regulation policies.
\medskip

We present LIME as a possible way of solving the issue in the CRM field. In fact, it has shown to be reliable on the majority of individuals we tested on, providing plausible explanations. However, as we reported herein, some weaknesses of the model raise suspicions whether the technique is yet enough mature to be considered as a standard solution.
\medskip

Possible improvements may comprise LIME’s theoretical advances and extended implementation, which may allow it to be integrated into the well-defined process of scorecards generation \cite{siddiqi_credit_2012}.
\medskip

We acknowledge financial support by CRIF S.p.A. and Università degli Studi di Bologna.

\printbibliography

\end{document}